\documentclass[runningheads]{llncs}

\usepackage{graphicx}
\usepackage{times}
\usepackage{soul}
\usepackage{url}
\usepackage[hidelinks]{hyperref}
\usepackage[utf8]{inputenc}
\usepackage[small]{caption}
\usepackage{graphicx}
\usepackage{amsmath}
\usepackage{booktabs}
\usepackage{subcaption}
\usepackage{floatrow}

\usepackage{todonotes}

\usepackage{algorithm}
\usepackage{algpseudocode}
\algtext*{EndIf}
\algtext*{EndProcedure}

\newcommand{\dom}{{\cal D}}
\newcommand{\mS}{{\cal S}}
\newcommand{\mV}{{\cal V}}
\newcommand{\mP}{{\cal P}}

\title{Compiling Stochastic Constraint Programs to And-Or Decision Diagrams}

\author{%
Behrouz Babaki\inst{1} \and
Golnoosh Farnadi\inst{1,2}  \and
Gilles Pesant\inst{1}
}

\institute{Polytechnique Montr\'{e}al \and Mila}

\begin{document}

\maketitle

\begin{abstract}
Factored stochastic constraint programming~(FSCP) is a formalism to represent multi-stage decision making problems under uncertainty. FSCP models support factorized probabilistic models and involve constraints over decision and random variables. These models have many applications in real-world problems. However, solving these problems requires evaluating the best course of action for each possible outcome of the random variables and hence is computationally challenging. FSCP problems often involve repeated subproblems which ideally should be solved once. In this paper we show how identifying and exploiting these identical subproblems can simplify solving them and leads to a compact representation of the solution. We compile an And-Or search tree to a compact decision diagram. Preliminary experiments show that our proposed method significantly improves the search efficiency by reducing the size of the problem and outperforms the existing methods.  
\end{abstract}

\section{Introduction}

Constraint satisfaction and optimization problems are usually assumed to be \emph{deterministic}, meaning that all parameters of the problem, also known as \emph{problem data}, are known with certainty. This ignores the complex, uncertain, and dynamic nature of the realworld problems.  Stochastic constraint programming is an attempt to address the problem of decision making under uncertainty using the constraint programming paradigm~\cite{Walsh02,hnich2011survey}. 

Recent developments in machine learning, together with abundance of collected data, has made it possible to capture our uncertain knowledge about the world as probabilistic models. \emph{Probabilistic graphical models} are a popular representation which assume a factorized joint distribution over random variables~\cite{Koller09}. This has motivated work on \emph{Factored Stochastic Constraint Programming}~(FSCP) which assumes that random variables follow such a factored model. FSCP allows us to model many applications in which decision variables are set before the random variables in alternating stages, i.e. we act first and observe later. For example, practical problems arise in transportation, finance, and the energy sector~\cite{wallace2005}. 

The state-of-the-art method for solving FSCP problems, called And-Or Branch and Bound~(AOBB), explores a search space consisting of two types of nodes: The \emph{And} nodes which correspond to random variables, and the \emph{Or} nodes which correspond to decision variables. To explore this search space efficiently, AOBB uses two pruning techniques that are commonly used in constraint satisfaction and optimization, namely constraint propagation and bounding~\cite{babaki2017stochastic}. However, these techniques are mainly applicable to the Or nodes. The presence of random variables calls for alternative techniques to improve search-space exploration. A similar issue has been encountered for search-based probabilistic inference algorithms, and has been addressed by identifying repeated subproblems, among other methods~\cite{BacchusDP09}. Identification of repeated subproblems has recently received attention in the constraint programming community, too~\cite{UnaGSS19,ChuBS12}. In this paper we apply this idea to the FSCP problems and demonstrate the gains that can be obtained from them in problems with repeated subproblems. The contributions of this work are:
\begin{itemize}
    \item Proposing a method for identification of repeated subproblems in FSCP problems,
    \item Compiling an And-Or search tree into a decision diagram,
    \item Extending a generic CP solver with the capability of performing And-Or search and compilation, and evaluating this approach through comparison with existing alternatives.
\end{itemize}

The paper is organized as follows. We first present the background material in Section~\ref{sec:background}. 
In Section~\ref{sec:scp} we present a brief description of FSCP and And-Or search. Section~\ref{sec:compilation} describes our method for caching the subproblems and compilation of FSCP into a decision diagram. We evaluate the proposed method in Section~\ref{sec:experiments}. We discuss the relation with existing work in Section~\ref{sec:related}, and conclude with directions for future research in Section~\ref{sec:conclusion}. 
\section{Background}
\label{sec:background}
In this section we review several key topics on which our proposed method relies. We start by reviewing multi-stage stochastic decision making problems and then review Bayesian networks, and decision diagrams.

\subsection{Multi-stage stochastic decision making}
\label{sec:dm}

We study a class of multi-stage stochastic decision making problems. At each stage of such problems the decision variables need to be set before the random variables are observed. In other words we act first and observe later. For example, we first need to decide how many workers we need to assign for a task and only later we observe the actual workload for the task. The goal is to assign values to decision variables at each stage in a way that the expected utility is maximized (or if desired, minimized). Note that in multi-stage stochastic problems the values chosen for the set of decision variables at each stage are conditioned both on the values of previously determined decision variables and the previously observed random variables. An example problem follows. 

\begin{example}[Production Planning]
\label{ex:walsh}
(from \cite{Walsh02}) 
In each quarter we sell between 101 and 105 items of a product. We need to satisfy the uncertain demand with probability 0.8 in every quarter. At the start of each quarter we decide how many books to print for the quarter, and the demand is known at the end of that quarter. The optimal production plan should minimize the expected cost of storing surplus items. 
\end{example}

The uncertainties of a multi-stage stochastic problem can be modeled as a factored distribution among the random and decision variables. Bayesian networks are one of the most popular representations of factored distribution that we review next.

\subsection{Bayesian Network}
\label{sec:bn}
A Bayesian network is a probabilistic graphical model which represents the conditional dependencies among a set of variables by edges in a directed graph. This representation facilitates compact encoding and efficient inference~\cite{Koller09}.
In addition to the graph structure
we must specify the conditional probability distribution at each node. If the variables are discrete, this can be represented as a \emph{conditional probability table}, which lists the probability that the child node takes on each of its different values for each combination of values of its parents. 
\begin{figure}
    \centering
    \includegraphics[scale=0.3]{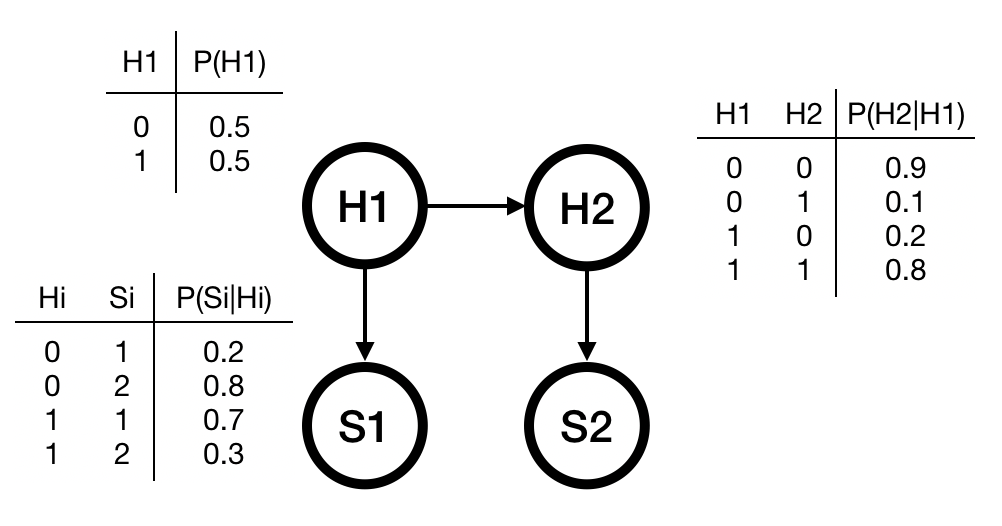}
    \caption{A Bayesian network representing a hidden Markov model. According to this model, the joint distribution $P(S_1, S_2, H_1, H_2)$ factorizes as $P(H1) \cdot P(S_1|H_1) \cdot P(H_2 | H_1) \cdot P(S_2 | H_2)$. }
    \label{fig:BookBN}
\end{figure}
A hidden Markov model which is a simple Bayesian network that captures a multi-stage stochastic process.
Figure~\ref{fig:BookBN} shows the structure and probability tables of a hidden Markov model.
\subsection{Decision Diagram}
\label{sec:decisiondiagram}

Decision diagrams are compact alternatives to decision trees. A decision diagram is a directed acyclic graph where nodes are variables and edges represent the assignment of value to the variables. Every path from a root node to a terminal node represents an assignment to all variables.

\section{Stochastic Constraint Programming}
\label{sec:scp}

Stochastic constraint programming (SCP) is a framework for modeling and solving multi-stage stochastic decision making problems. A multi-stage stochastic constraint satisfaction program is defined as a 7-tuple $\mathbf{P} = \langle
\mV, \mS, \mathcal{D}, \mathcal{P}, \mathcal{C}, \theta, \prec
\rangle$~\cite{hnich2011survey}. $\mV$ and $\mS$ are decision variables and random (stochastic) variables, respectively. $\mathcal{D}$ is the domain of variables in $\mV \cup \mS$, $\mathcal{P}$ is a function that for each variable in $\mS$ defines a probability distribution over its domain. $\mathcal{C}$ is a set of constraints. Each constraint is specified over a non-empty subset of $\mV$ and a (possibly empty) subset of $\mS$.  
$\theta$ is a function that assigns a minimum satisfaction probability to each constraint in $\mathcal{C}$.
$\prec$ is a partial ordering over $\mV \cup \mS$: $\mV_1 \prec \mS_1 \prec \ldots \prec \mV_T \prec \mS_T$. The sets $\mV_i$ and $ \mS_i$ respectively partition $\mV$ and $\mS$  and can be possibly empty, and $T$ is the number of stages.  

A solution to the stochastic constraint program is a \textit{policy tree} where each path represents an assignment to the variables in $\mV \cup \mS$, and follows the ordering $\prec$. 
In this tree each decision variable has just one child (corresponding to the selected value) and each random variable has as many children as the number of values in its domain. For each constraint in ${\cal C}$, the sum of probabilities of paths in which the constraint is satisfied should meet the minimum probability requirement specified by $\theta$. 

Given a \emph{utility} function $U(\mV, \mS)$ this definition can be extended to an optimization setting where the objective is to maximize (or minimize) the expected utility:
\begin{align}
\max_{\mV_1} \sum_{\mS_1} \ldots \max_{\mV_T} \sum_{\mS_T} P(\mS) \times U(\mV, \mS) \label{eq:exputil}
\end{align}

\subsection{Factored Stochastic Constraint Program}

The assumption of independent random variables falls short of representing the existing correlations between random variables in the real world. This motivates a generalization in which $\mP$ specifies a join probability distribution over variables in $\mS$. In factored stochastic constraint programming (FSCP) the join distribution is factorized, i.e. $P(\mathcal{S}) = \prod_{\mathcal{S}_i \subset \mathcal{S}} \phi(\mathcal{S}_i)$. This is the assumption that we are making in the rest of this work. We also make the same extra assumptions as those made by ~\cite{babaki2017stochastic}: 1) The utility function is represented by a single \emph{utility variable}. This is not a restriction as long as the utility function can be encoded by a set of constraints. 2) The threshold assigned to all constraints by $\theta$ is one, i.e. all constraints are \emph{hard} and should be satisfied in all \emph{possible} (non-0 probability) paths.

\begin{figure}[h]
    \centering
    \floatbox[{\capbeside\thisfloatsetup{capbesideposition={right,center},capbesidewidth=3.5cm}}]{figure}[\FBwidth]
    {\caption{The optimal policy tree of Example~\ref{ex:book}. The decision in the second quarter ($V_2$) depends on the realized demand in the first quarter ($S_1$). }\label{fig:policy}}
    {\includegraphics[scale=0.15]{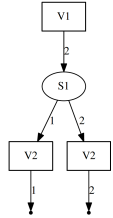}}
\end{figure}

\begin{example}[Production Planning, continued]
\label{ex:book}
Assume that in Example~\ref{ex:walsh} the demand and supply in quarter $i$ are represented respectively by random variable $S_i$ and decision variable $V_i$, both with domain $\{1, 2\}$. The demand depends on the market sentiment (represented by random variable $H_i$), which itself depends on the market sentiment in the previous quarter. The goal is to minimize the expected number of unsold books in the last quarter, while disallowing shortages.
\end{example}
Assuming $T=2$ quarters, the dependencies between random variables can be represented by the Bayesian network of Figure~\ref{fig:BookBN}. The objective function, constraints, and domains of the corresponding FSCP are as follows:
{\small
\begin{align}
    &\min_{V_1} \sum_{S_1} \min_{V_2} \sum_{S_2} P(S_1, S_2)  \times U  \nonumber \\
    &\text{s.t.}  \nonumber \\
    & W = V_1 - S_1  \label{cons:l1} \\
    & U = W + V_2 - S_2  \label{cons:l2} \\
    & V_1, S_1, V_2, S_2 \in \{1, 2\}  \quad W \in \{0,\ldots,2\}  \quad U \in \{0, \ldots, 4\}  \nonumber  
\end{align}
}
The optimal policy tree for Example~\ref{ex:book} is shown in Figure~\ref{fig:policy}.

The structure of an FSCP problem can be summarized in a graphical representation called the \emph{factor graph}. We will later use this structure the factor graph to identify the identical subproblems during search. 

\begin{figure}
    \centering
    \includegraphics[scale=0.22]{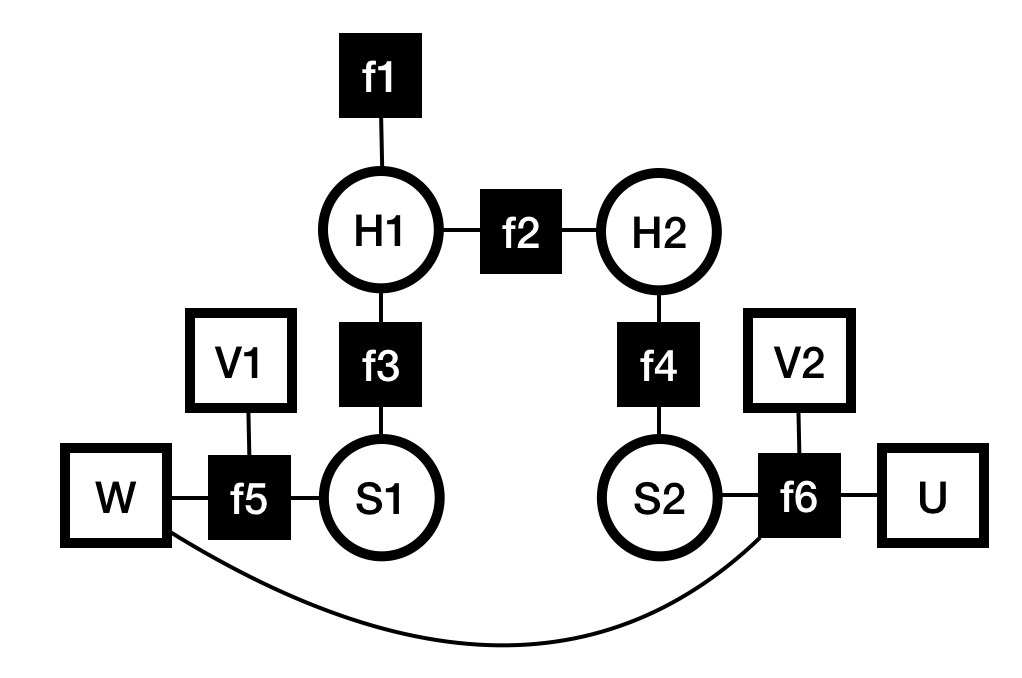}
    \caption{The factor graph of Example~\ref{ex:book}. White squares are variables and black squares represent the factors. Factors represented by $f_1, \ldots, f_4$ correspond to conditional probabilities from the Bayesian network of Figure~\ref{fig:BookBN}. Factors $f_5$ and $f_6$ correspond to the constraints in Equations~\ref{cons:l1} and~\ref{cons:l2}.}
    \label{fig:factor}
\end{figure}

\begin{definition}[factor graph]
The factor graph is a bipartite graph which represents the factorization of a function with several variables.
An FSCP factor graph can be represented by a graph $G=(V \cup F, E)$ where each $v \in V$ corresponds to a variable, and each node $f \in F$ corresponds to a \emph{factor} (that is, a constraint or conditional probability table). The nodes $v \in V$ and $f \in F$ are connected to each other if and only if the variable corresponding to $v$ appears in the scope of the factor corresponding to $f$. 
\end{definition}
The factor graph of Example~\ref{ex:book} is shown in Figure~\ref{fig:factor}. 

\subsection{Solving FSCP using And-Or search}

The expression in Equation~\ref{eq:exputil} can be represented by a graphical structure called the \emph{And-Or search tree}. Solving an SCP problem, i.e. evaluating this expression, is possible by traversing this tree. An And-Or search tree has two types of internal nodes: 1) \emph{And} nodes which correspond to random variables and \emph{sum} operator, and 2) \emph{Or} nodes  which correspond to decision variables and \emph{max} operator. An edge represents the assignment of a value to the variable that corresponds to the source node. A path from the root to a leaf represents an assignment to every variable in $\mV \cup \mS$ and the order of variables on each path follows $\prec$. 

Given an assignment $(\mV = v, \mS = s)$ on a path, the \emph{value} of the leaf node is defined as $P(s)U(v,s)$. The value of an internal node $u$ can be computed recursively: If $u$ corresponds a random variable, $value(u) = \sum_{c \in \text{children}(u)} value(u)$. Otherwise $value(u) = \max_{c \in \text{children}(u)} value(u)$. The optimal policy can be extracted by examining the trace of a bottom-up traversal of the tree. 

\begin{figure*}
    \centering
        \includegraphics[scale=0.14]{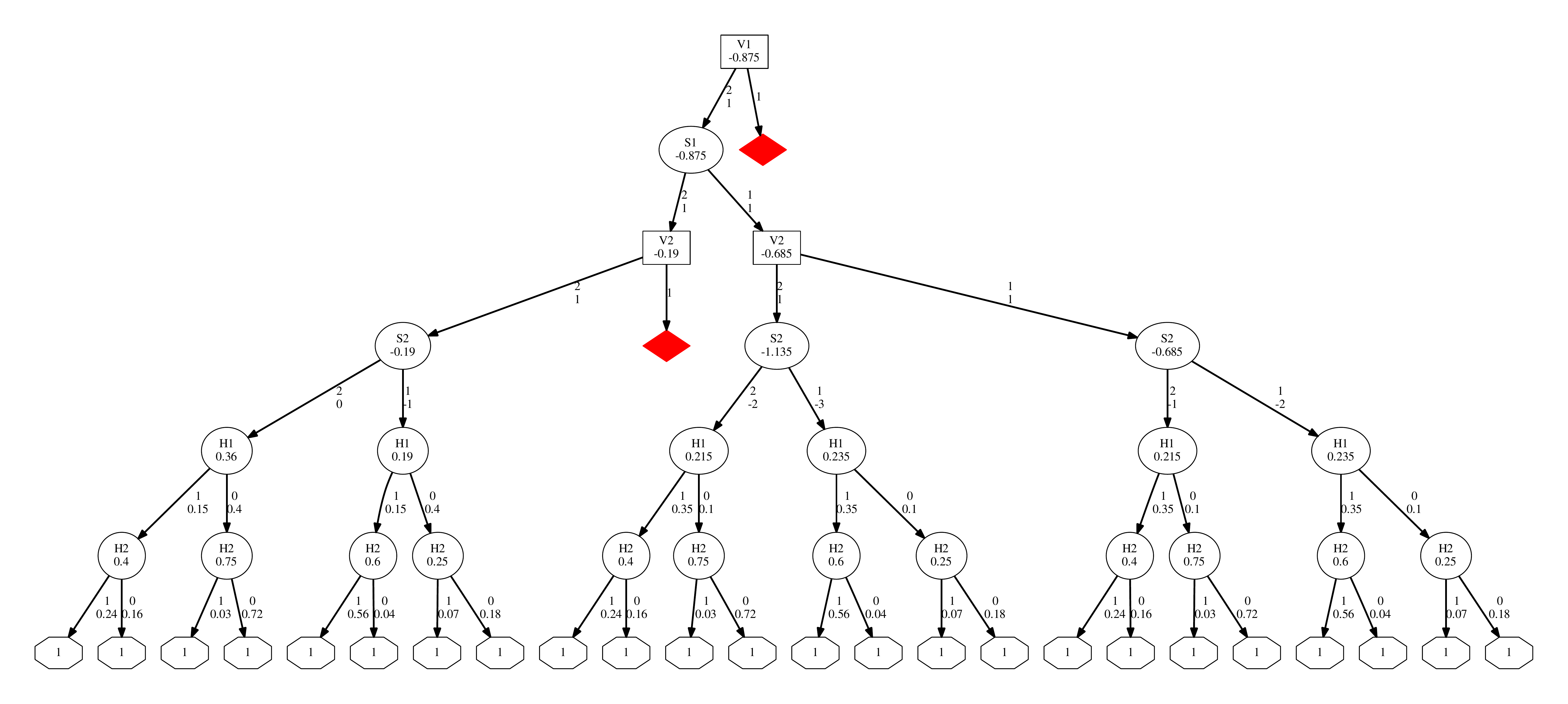}%
        \includegraphics[scale=0.14]{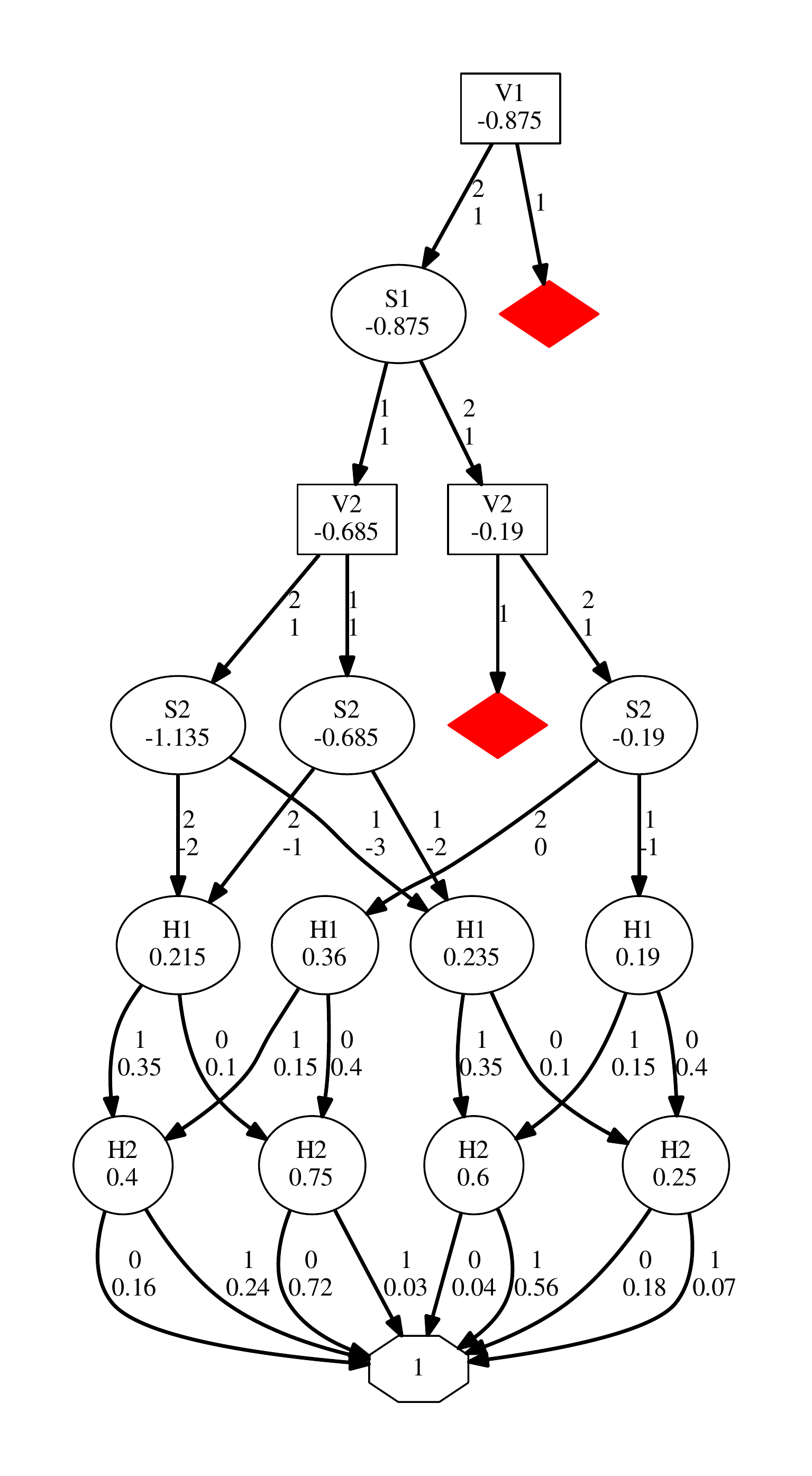}
    \caption{The And-Or search tree (left), and And-Or decision diagram (right) for Example~\ref{ex:book}. The edges are annotated with two values: the assignment to the source variable (top), and the edge weight (bottom). The nodes are annotated with the name of corresponding variable (top) and the node value (bottom). Failure is represented by a red diamond.}
    \label{fig:trees}
\end{figure*}

Instead of storing the value $P(s)U(v,s)$ at the leaves, we can take advantage of the factorization of probabilities and store these values on edges of the tree. Recall that an edge represents the assignment of a value to a variable. This assignment might reduce some factors of the distribution to a value. It might also reduce the domain of the utility variable to a value. We define the weight of an edge as the product of these values. Figure~\ref{fig:trees} (left) shows the And-Or search tree of Example~\ref{ex:book}. 

When constraints are present, only subtrees that satisfy the constraints are included in this evaluation procedure. In such cases we can use constraint propagation to explore the search space more efficiently. Since the hard constraints should be satisfied in all possible scenarios, two modifications should be made to the standard constraint programming machinery: First, failure of any child of an And node immediately fails the node itself. Second, a reduction in the domain of a random variable caused by propagation is considered failure, as it implies that there is a possible scenario in which a constraint is violated.

The procedure described above uses constraint reasoning to prune the search space. Another possible improvement is to establish bounds that will guarantee that a subtree cannot lead to a solution better than what is already obtained. The And-Or Branch-and-Bound method uses such bounds to further prune the search space~\cite{babaki2017stochastic}. 

\section{Compiling FSCP to Decision Diagram}
\label{sec:compilation}

Processing a factorized model can sometimes result in solving identical subproblems repeatedly. Some search-based algorithms for processing graphical models avoid these redundant computations by identifying identical subtrees in the search tree and merging them, hence obtaining a compact equivalent graph~\cite{MateescuDM08}. 

The factorized nature of FSCP problems suggests the possibility of applying a similar approach to these problems. This will turn the search tree into a graph which we call an \emph{And-Or Decision Diagram (AODD)}. However, merging identical subtrees repeatedly is not a practical method for compiling FSCPs, as it requires construction of the And-Or search tree. In this section, we describe a method for generating AODDs during the search, without the need to materialize the full search tree. 

We traverse the And-Or search tree in a depth-first manner. However, before expanding each node, we first check whether the subtree rooted at this node is identical to another subtree visited earlier during the search. If this is the case, instead of expanding the node we connect its parent to the root of the existing subtree. 

The described procedure depends on a method for testing the equivalence of subproblems without exploring them. Each subtree is uniquely identified by assignment to the variables preceding this node in the tree. However, it can be the case that the subproblem only depends on a subset of those variable. Following the terminology used in the probabilistic reasoning community, we call this subset the \emph{context} of the subproblem: 

\begin{definition}[context]
 For every internal node in the And-Or search tree, the path from the root to that node defines a (partial) assignment. We call a factor (i.e. constraint or probabilistic factor) \emph{active} if it has some unassigned variable in its scope. The \emph{context} of a node is the set of assignments to variables on its path which are in the scope of some active factor. 
\end{definition}

In Example~\ref{ex:book}, variable $H_1$ is assigned after $S_1, V_1, S_2, V_2$. Figure~\ref{fig:BookBN} shows that the context of this variable is $\{S_1, S_2\}$. As one can observe in Figure~\ref{fig:trees}, the subtrees for assignments $\{V_1=2, S_1=1, V_2=2, S_2=2\}$ and $\{V_1=2, S_1=1, V_2=1, S_2=2\}$ are identical and can be merged. A similar case holds for subtrees of assignments $\{V_1=2, S_1=1, V_2=2, S_2=1\}$ and $\{V_1=2, S_1=1, V_2=1, S_2=1\}$.

Algorithm~\ref{alg:compile} summarizes the procedure for compiling an FSCP over domain $\dom'$ into an AODD. Before solving each subproblem, the cache key is generated from the context of the subproblem root node (Line~\ref{aodd:key}). The cache is then inspected ( Line~\ref{aodd:inspect}). If an identical subproblem is found, the node is merged with the existing subgraph. Otherwise, the search proceeds. The value of each node is calculated based on the values of its children (Line~\ref{aodd:val_begin}-\ref{aodd:val_end}), and the node is stored in the cache before backtracking (Line~\ref{aodd:store}).

\begin{algorithm}[t]
\begin{algorithmic}[1]
\Procedure{\textsc{aodd}}{$\dom'$}
  \If{\textsc{propagate}($\dom'$) == \text{failure}}
    \State{{\bf return} \text{failure}}
  \EndIf  
  \If{$\forall x \in {\cal V} \cup {\cal S}: |\dom'(x)| = 1$}
    \State{{\bf return} leaf node}
  \EndIf
  \State {$\psi \gets \textsc{context}(\dom')$} \label{aodd:key}
  \If{$\textsc{cache}(\psi) \neq \textsf{nil}$} \label{aodd:inspect}
    \State{{\bf return} $\textsc{cache}(\psi)$} 
  \EndIf
  \State{Select unassigned variable $X$ according to $\prec$}
  \State{Create new node $u$}
  \For{$x \in \dom'(X)$}
    \State{$\dom'' \gets \dom'$}
    \State{$\dom''(X) \gets \{x\}$}
    \State{$w \gets \textsc{ComputeEdgeWeight}(\dom', \dom'')$}
    \State{$v \gets \textsc{aodd}(\dom'')$}
    \If{$v$ == \text{failure} and $X \in \mathcal{S}$}
        \State{\textbf{return} \text{failure}}
    \EndIf
    \State{Connect node $u$ to $v$ with weight $w$}
    \If{$v$ is the first child} \label{aodd:val_begin}
        \State{$value(u) \gets value(v)$}
    \Else
        \If{$X \in \mathcal{S}$}
            \State{$value(u) \gets value(u) + value(v)$}
        \Else
            \State{$value(u) \gets \max\big(value(u), value(v)\big)$} \label{aodd:val_end}
        \EndIf
    \EndIf
  \EndFor
  \If{$u$ has no child}
     \State{$u \gets \text{failure}$}
  \EndIf
  \State{$\textsc{cache}(\psi) \gets u$} \label{aodd:store}
  \State{{\bf return} $u$}
\EndProcedure
\end{algorithmic}

\caption{Compiling the And-Or decision diagram \label{alg:compile}}
\end{algorithm}

Figure~\ref{fig:trees} (right) shows the And-Or decision diagram of Example~\ref{ex:book} obtained using the described method. It can be observed that the identical subproblems which we mentioned earlier, are now merged. Once the AODD is generated, the optimal policy can be retrieved in the same way that it is obtained from an And-Or search tree. 

The proposed method makes it possible to compile an FSCP to AODD on the fly, by introducing a small modification to the And-Or search procedure. This can lead to significant performance gains, as demonstrated in the next section. 

\section{Experiments}
\label{sec:experiments}

To evaluate our approach, in this section we investigate the following research questions:
\begin{itemize}
    \item [\textbf{Q1}] How effective is our method in identifying identical subproblems and compressing the search tree?
    \item [\textbf{Q2}] What is the effect of compilation on the performance of And-Or search compared with the existing methods?
    \item [\textbf{Q3}] When doesn't the compilation help? 
\end{itemize}

We address the above research questions using the \emph{knapsack} and the \emph{investment} problems. \begin{itemize}
    \item {\bf Knapsack} problem (based on a problem from 
    \cite{hnich2011survey}): Consider a knapsack with a certain capacity. Assume at each stage, an item has arrived and we need to choose to pick the item or leave it. The weight and value of each item is stochastic and is observed after making the decision. There are 5 different possibilities for items' weight, and 3 possibilities for their values. The goal is to maximize the expected sum of values of the collected items subject to the hard constraint that the total weight of the items is less than the capacity of the knapsack. We implemented two variations of this problem: In the \emph{independent} version (Knapsack-I), all variables are independent and in the \emph{chain} version (Knapsack-C), weight and value at each stage depend on the similar variables at the previous stage.

    \item {\bf Investment} problem (based on a problem from \cite{babaki2017stochastic}): Consider a company that has two options for investment at the start of each season, and only at the end of the season observes the stochastic return. 
    There are 4 possibilities of return for each investment option. 
    The first option has a higher return on average but the second option brings more tax relaxation at the end of the horizon. The goal is to maximize the expected returns by considering the tax relaxations. Similar to the previous problem, we have two variations of this problem (denoted by Investment-I and Investment-C). Note that for this problem, we have a hard constraint that the total sum of return of the second option should be less or equal to the total sum of return of the first option.
\end{itemize}

We ran experiments on machines with an Intel i5-4590 processor (3.3GHz) and 8GB of RAM running Linux Ubuntu 16.04. We extended the constraint programming solver Mini-CP~\cite{minicp} with And-Or search, caching, and compilation functionality. The time-out used was 1800 seconds. The MIP solver is Gurobi-8.1\footnote{\url{www.gurobi.com}}. 

\begin{figure}[h]
    \centering
    \includegraphics[width=0.65\textwidth]{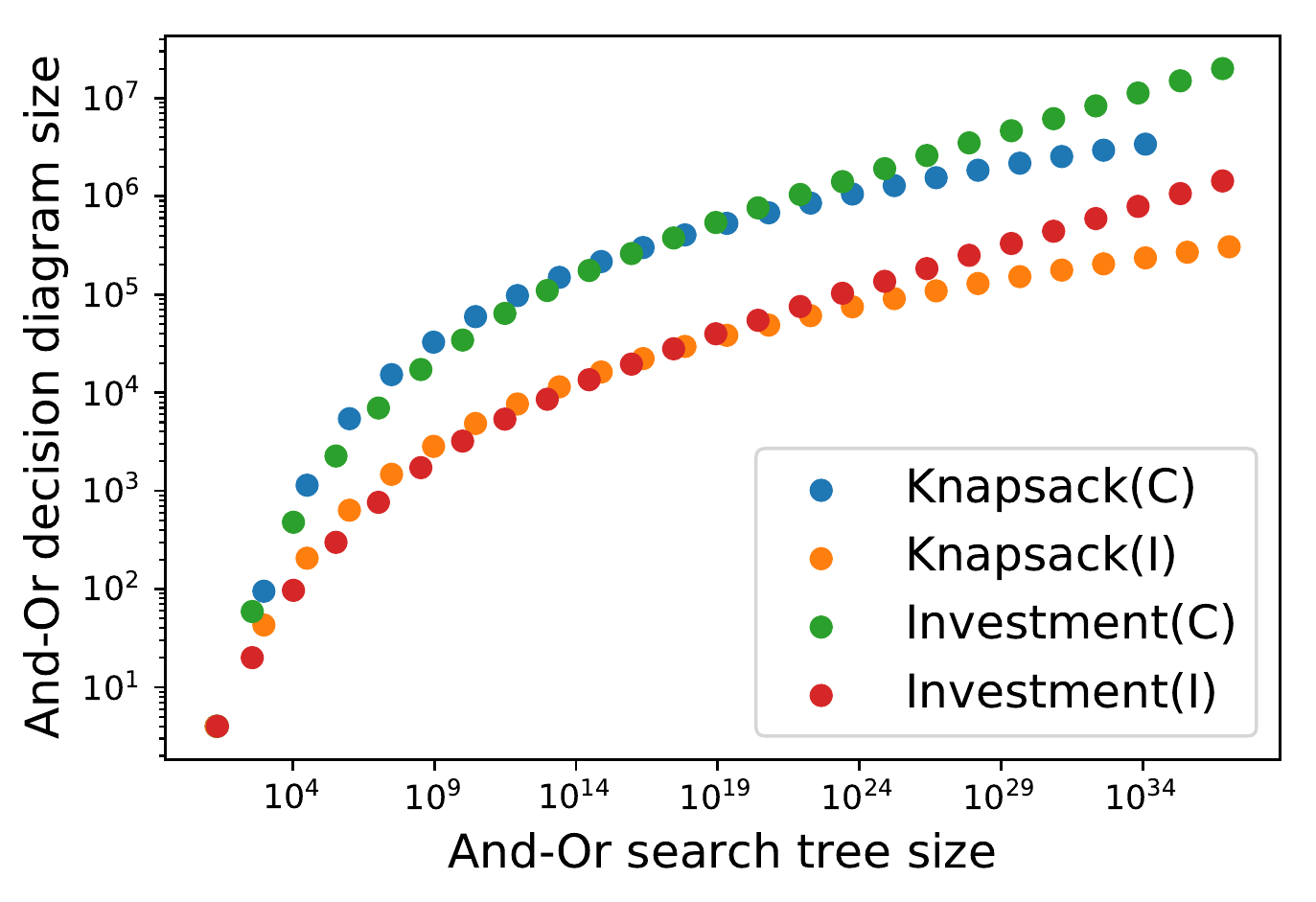}
    \caption{Comparing the size of And-Or search tree and decision diagram. Note that the scales are logarithmic.}
    \label{fig:reduction}
\end{figure}

To address \textit{Q1}, we compare the performance of our approach on both problems with and without compilation. 
We measure the effects of compilation by varying the number of stages in both problems.

As shown in Figure~\ref{fig:reduction}, compilation leads to significant reductions for both problems. As the number of stages increase, so does the number of identical subproblems, which in turn results in exponential reductions. 

\begin{figure}[h]
    \centering
    \begin{subfigure}[t]{0.5\textwidth}
        \centering
        \includegraphics[width=\textwidth]{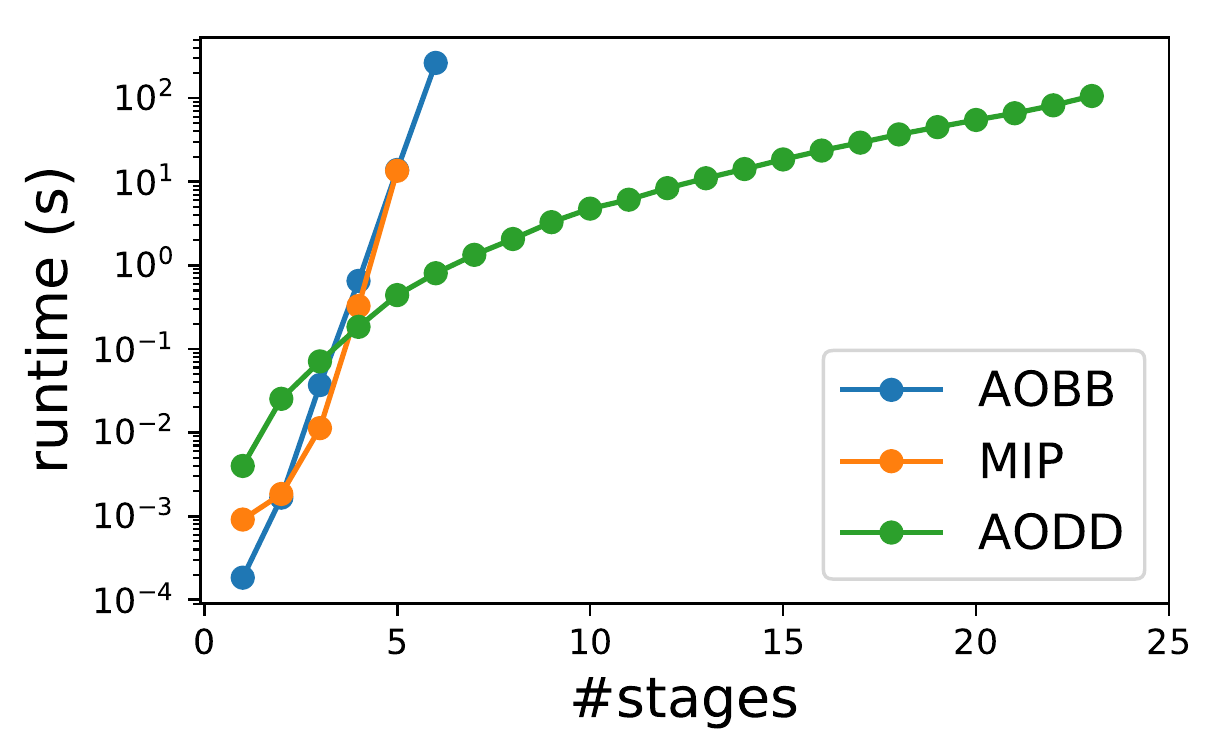}
        \caption{Knapsack-C}
    \end{subfigure}%
    ~ 
    \begin{subfigure}[t]{0.5\textwidth}
        \centering
        \includegraphics[width=\textwidth]{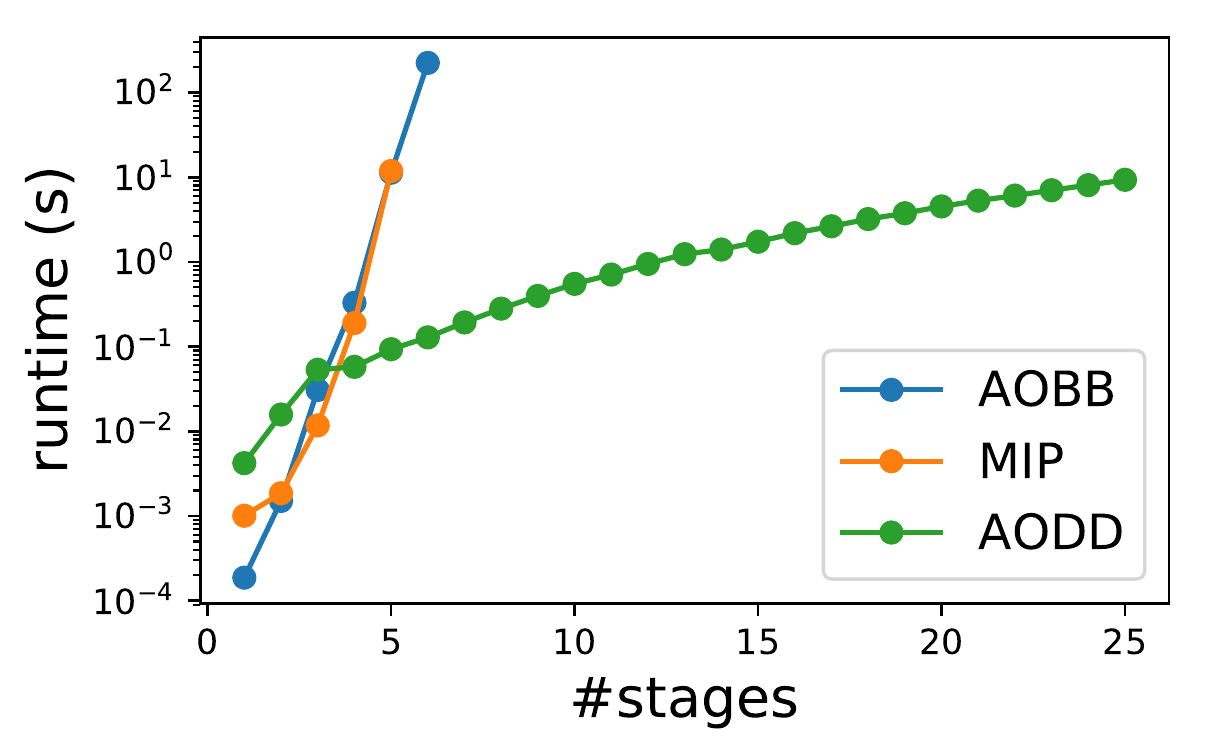}
        \caption{Knapsack-I}
    \end{subfigure}\\
    ~
    \begin{subfigure}[b]{0.5\textwidth}
        \centering
        \includegraphics[width=\textwidth]{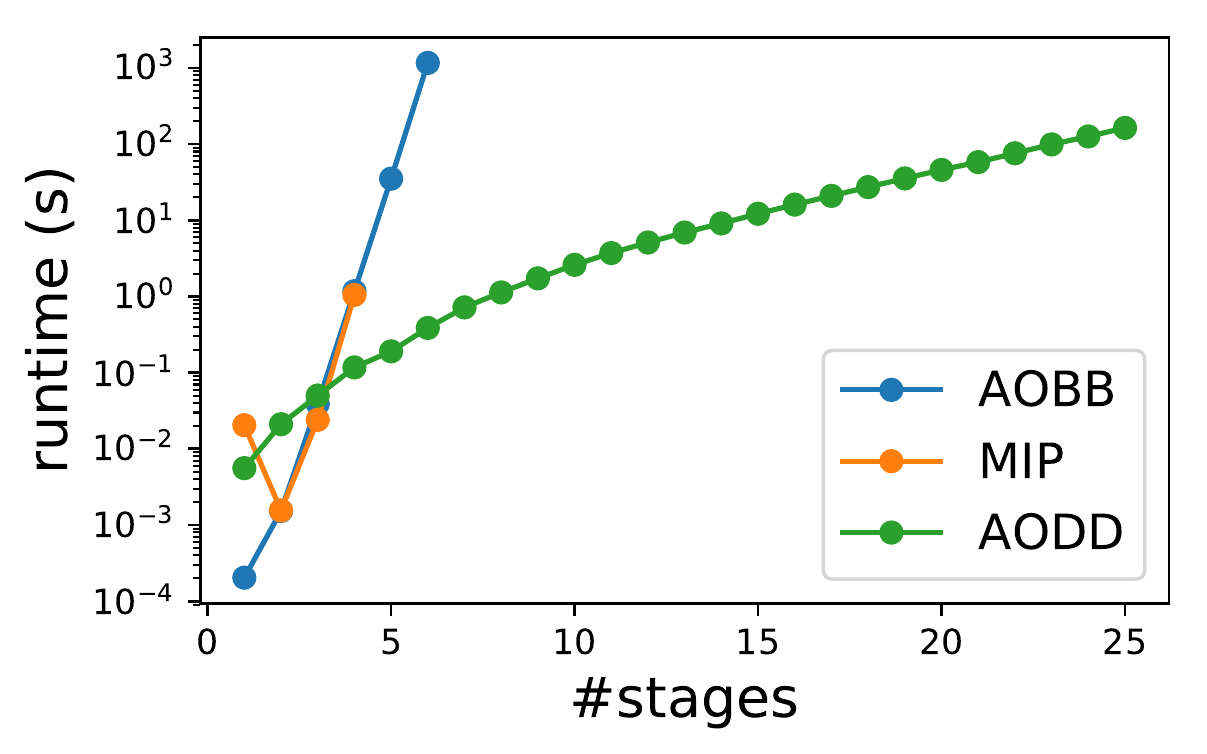}
        \caption{Investment-C}
    \end{subfigure}%
    ~
    \begin{subfigure}[b]{0.5\textwidth}
        \centering
        \includegraphics[width=\textwidth]{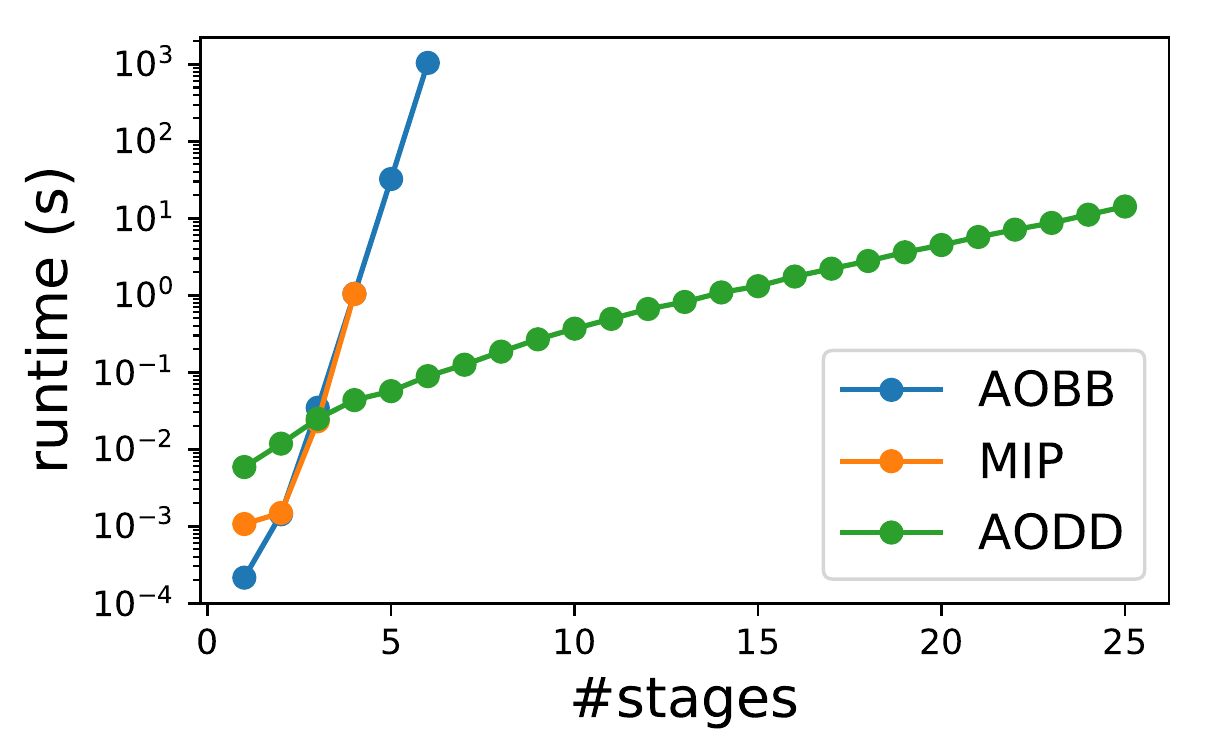}
        \caption{Investment-I}
    \end{subfigure}
    \caption{Comparing the runtime of our method (AODD) with scenario-based (MIP) and And-Or branch and bound (AOBB) approaches. The runtime is measured in seconds.}
    \label{fig:runtime}
\end{figure}

To investigate the performance of our approach compared with the existing methods and address \textit{Q2}, we compare our algorithm with the scenario-based conversion to MIP and And-Or branch and bound approach~(AOBB)~\cite{babaki2017stochastic}. The results presented in Figure~\ref{fig:runtime} show that without compilation, we are not able to solve problems beyond 6 stages using both MIP and AOBB approaches. However our approach scales to 25 stages and easily solves these problem in less than 5 minutes. 

\begin{table}[t]
    \centering
    \begin{tabular}{c|c c c}
    \toprule
       \# stages & MIP & AOBB & AODD \\
    \midrule
        1 & 0.0240 & 0.0002 & 0.0067\\
        2 & 0.0019 & 0.0016 & 0.0409\\
        3 & 0.0116 & 0.0351 & 0.1965\\
        4 & 0.3221 & 0.6099 & 2.7660\\
        5 & 3.4789 & 3.9858 & 58.4728\\
        6 & M & 76.6795 & M \\
        7 & M & T & M \\
    \bottomrule
    \end{tabular}
    \caption{Comparing the runtime (s) of our method AODD with AOBB and MIP using the \emph{Knapsack(H)} problem. We either run out of memory (M) during generation of the problem or timeout (T) when solving the problem. 
    }
    \label{tab:nohelp}
\end{table}

\begin{figure}[h!]
    \centering
    \includegraphics[width=0.65\textwidth]{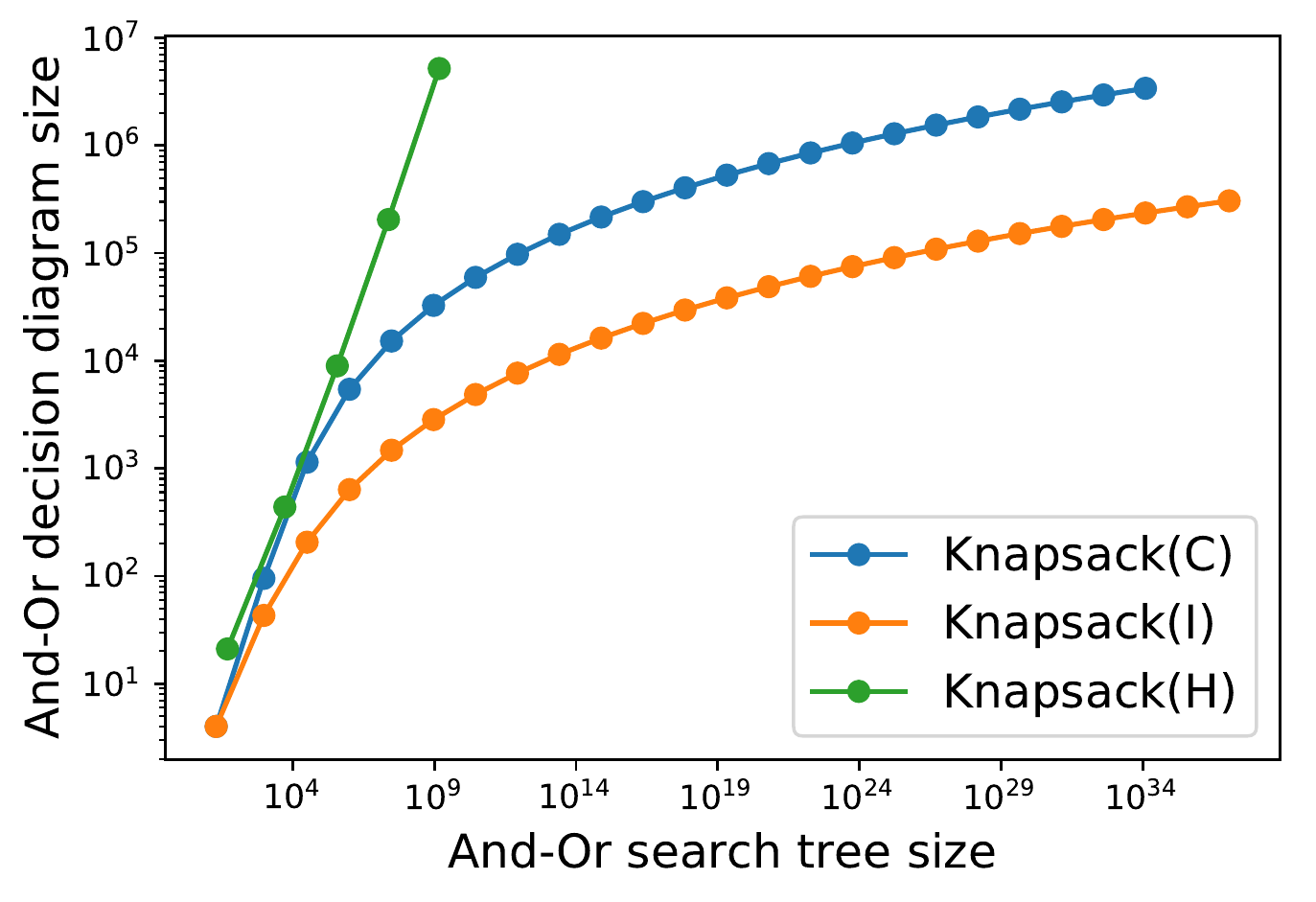}
    \caption{Comparing the And-Or search tree and decision diagram size of different types of the Knapsack problem using the compilation approach. Both scales are logarithmic.}
    \label{fig:hmm}
\end{figure}

It is important to note that the compilation is more effective when the number of identical subproblems is high. 
Hence, the structure of the model affects its performance. To address \textit{Q3}, we consider the knapsack problem and change the Bayesian network by including a hidden variable per item as described in~\cite{babaki2017stochastic}. We refer to this model as \emph{knapsack-H}. 
The results in Table~\ref{tab:nohelp} show that compilation is less effective for this variant of the knapsack problem compared to the chain and independent versions. 
While solving 25-stages Knapsack-C takes only 9 seconds and 20-stages Knapsack-I takes 1 minutes to solve using AODD, Knapsack-H is not solved beyond 5 stages. 
When solving Knapsack-H, all hidden variables appear last in the ordering. Since all other variables depend on these hidden variables, there are no identical subproblems before reaching the hidden variables in the search tree. This leads to less reduction in Knapsack-H compared to the other two variants (see Figure~\ref{fig:hmm}).
The AOBB approach takes advantage of bounding to solve the 6-stages problem, which suggests a future direction to explore bounding in AODD.

\section{Related work}
\label{sec:related}

Our method is closely related to the And-Or search trees for graphical models (for example see~\cite{MateescuDM08}). In those studies the And-Or nodes have a different meaning from ours, where an And node corresponds to problem decomposition, and an Or node represents branching. Most of these works assume only one type of variable (only decision or random variables). \emph{Mixed deterministic-probabilistic networks}~\cite{MateescuD08} include both deterministic and probabilistic factors, but only include decision variables, and solve the probabilistic reasoning problems (e.g. MPE and MAP inference) subject to constraints. 
To the best of our knowledge, \cite{Marinescu09} is the only work in this area which includes both decision and random variables. This work evaluates influence diagrams using And-Or search graphs and uses a SAT solver to avoid exploring the subproblems with zero probability. Our method generalizes this approach by incorporating hard constraints on decision variables and using  global constraints and the propagation power of CP.

Factored SCPs bridge the gap between influence diagrams and stochastic constraint programming by imposing probabilistic and deterministic factors over decision and random variables~\cite{babaki2017stochastic}. And-Or search with branch and bound has been previously used to evaluate influence diagrams when no constraint propagation is involved~\cite{YuanWH10}. And-Or search trees have also been used to solve stochastic constraint programs with independent random variables~\cite{Walsh02}. Neither of these methods exploits identical subproblems during the search.    

Compilation to decision diagrams is a well-known technique in AI with celebrated success in model counting~\cite{MuiseMBH12}, probabilistic inference~\cite{ChoiKD13}, probabilistic logic programming~\cite{FierensBRSGTJR15}, and planning~\cite{SpeckGM18,HoeySHB99}, among others. Recently, decision diagrams have received attention in combinatorial optimization, too. However, in most of these studies the construction of decision diagrams is problem-specific. A notable exception is the work of~\cite{UnaGSS19} which proposes methods for compilation of CP subproblems to decision diagrams. This study proposes sophisticated methods for identification of identical subproblems, which require interaction with the propagation algorithms. 

Scenario-based approaches are approximate methods that solve SCP problems by sampling a subset of possible scenarios from the probability distribution~\cite{TarimMW06,HemmiT018}. Our work demonstrates that it is possible to take all scenarios into account without the need to explicitly enumerate them. 

\section{Conclusion \& Future Work}
\label{sec:conclusion}

We presented a method for compiling factored stochastic constraint programs into And-Or decision diagrams. Our experiments demonstrate the advantages of such a compilation, especially when there is a lot of redundancy in the search space. 

Decision diagrams have been successfully used in combinatorial optimization for obtaining bounds~\cite{BergmanCHH16}. A direction for future work is to devise compilation methods that create relaxed And-Or decision diagrams, which can then be used for obtaining bounds during search. There exists some recent work on using more sophisticated techniques for identification of equivalent and dominant subproblems~\cite{UnaGSS19,ChuBS12}. This motivates future work on subproblem identification in And-Or search. 

The SCP problems usually include \emph{chance} constraints, i.e. constraints that should be satisfied at least in a certain fraction of possible scenarios. Our current formalism only considers hard constraints. Generalizing this work to include chance constraints is another promising direction for future work. 

\bibliographystyle{splncs04}
\bibliography{references}

\end{document}